\documentclass[11pt]{article}
\usepackage[final]{acl}

\usepackage{times}
\usepackage[raster,skins,most]{tcolorbox} 
\usepackage{multirow}
\usepackage[table]{xcolor}
\definecolor{lightblue}{rgb}{0.9, 0.95, 1.0}

\usepackage{CJKutf8}
\usepackage[T1]{fontenc}
\usepackage[utf8]{inputenc}

\usepackage{latexsym}
\usepackage{algorithm}
\usepackage{algpseudocode}
\usepackage{amsmath}
\usepackage{amssymb}
\usepackage{booktabs}
\usepackage{microtype}
\usepackage{inconsolata}
\usepackage{graphicx}

\tcbuselibrary{breakable,skins}
\colorlet{red1}{red!80}

\title{SlimSearcher: Training Efficiency-Aware Web Agents \\via Adaptive Reward Gating} 
\author{
  \textbf{Zequn Xie\textsuperscript{*}},
  \textbf{Junjie Wang\textsuperscript{*}},
  \textbf{Dan Yang},
  \textbf{Jie Feng},\\
  \textbf{Yue Shen},
  \textbf{Jian Wang},
  \textbf{Jinjie Gu}\\[6pt]
  \textsuperscript{1} Zhejiang University,
  \textsuperscript{2} Ant Group \\[4pt]
  {\small \textbf{Correspondence:} 
    \texttt{zqxie@zju.edu.cn}, 
  \texttt{wangjj2018@zju.edu.cn}, 
}}

\begin{document}
\maketitle

\begin{abstract}
Deep research agents have demonstrated remarkable capabilities in complex information-seeking tasks, yet this power comes at a steep computational cost. Driven by accuracy-focused training paradigms, current models adopt brute-force strategies characterized by \textit{blind tool dependency} and \textit{performative reasoning}---generating long, redundant trajectories that are far from necessary for resolving these tasks, leading to wasteful tool calls and excessive token consumption. To overcome this efficiency trap, we propose \textbf{SlimSearcher}, a principled framework that pushes the Pareto frontier between accuracy and computational cost across both Supervised Fine-Tuning (SFT) and Reinforcement Learning (RL). In the SFT stage, SlimSearcher employs Pareto-efficient filtration to distill trajectories that are both successful and economical, guiding the model toward inherently efficiency-aware search behaviors. During RL, we introduce \textit{Adaptive Reward Gating}, a dynamic reward-shaping mechanism that evaluates relative tool and token efficiency within a sampled cohort. By cascading these adaptive efficiency metrics with a strict correctness gate, our approach effectively avoids the brevity bias associated with absolute penalties and mitigates reward hacking. Extensive experiments on long-horizon benchmarks, including GAIA, BrowseComp, and XBench-DeepSearch, demonstrate that SlimSearcher reduces average tool-call rounds by 17\%--58\% while maintaining or improving accuracy.Our code is available at: \url{ https://github.com/AQ-MedAI/AntAFu-DeepResearch}
\end{abstract}
\begin{figure}[htbp]
    \centering
    
    \includegraphics[width=\linewidth]{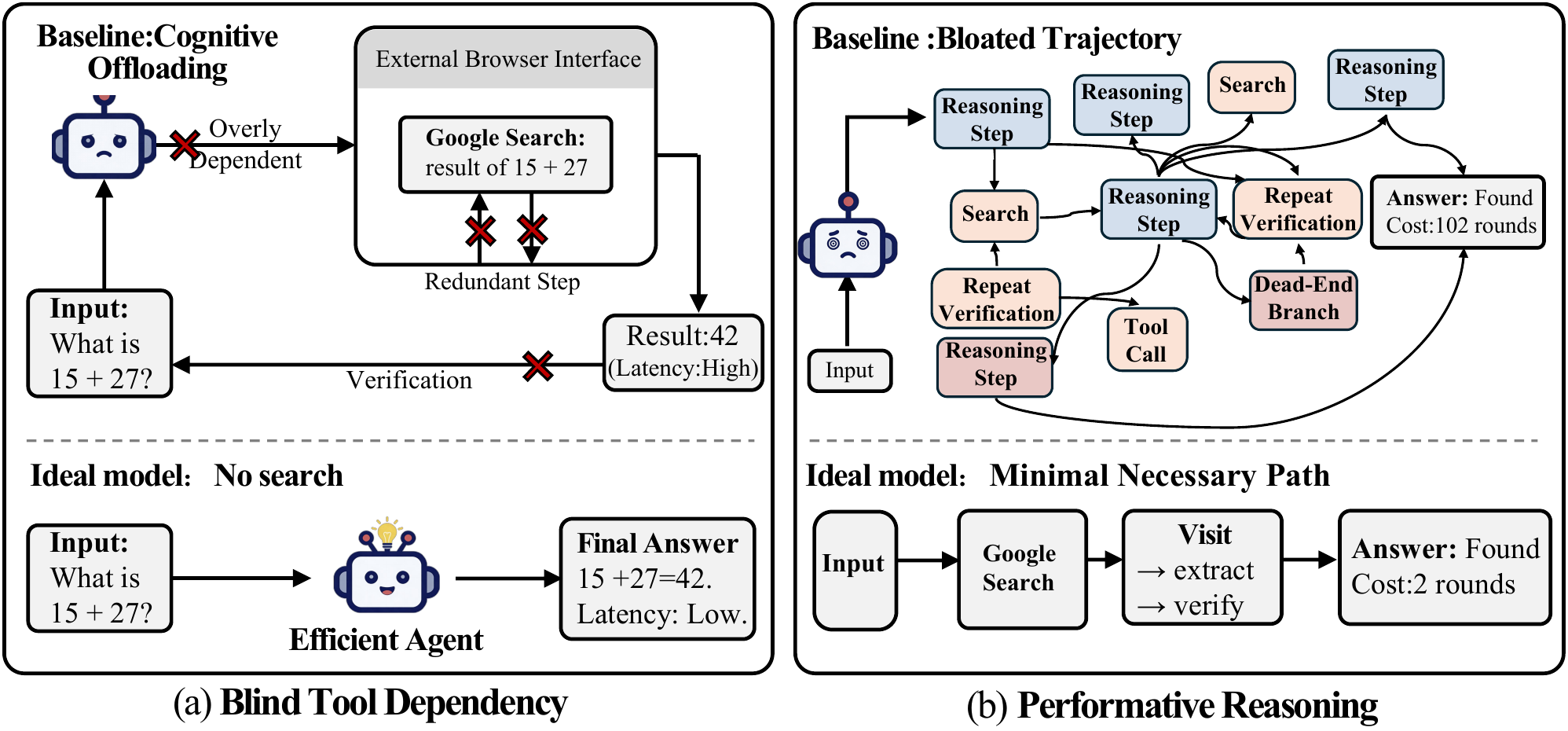}
    
\caption{\textbf{Behavioral Analysis of the Efficiency Trap.} 
(a) \textbf{Blind Tool Dependency}: The baseline agent indiscriminately invokes external search tools for a common-sense query resolvable via internal knowledge, leading to increased latency. 
(b) \textbf{Performative Reasoning}: As demonstrated in the complex query case (see Figure \ref{fig:xbench_task6_comparison}), even in a successful trajectory, the baseline agent generates redundant loops and dead-end branches. 
}
    \label{fig:motivation_behavior}
\end{figure}

\section{Introduction}

With the rapid advancement of artificial intelligence, LLMs have progressed from static text generators to sophisticated agents that can leverage external tools and interact with dynamic environments \cite{kimik2,glm45}. By integrating language understanding with external tools such as search engines and code interpreters, tool-integrated reasoning substantially expands the problem-solving capabilities of LLMs beyond purely linguistic reasoning. This capability not only improves adaptability in open-ended tasks but also supports a new generation of powerful systems, ranging from commercial offerings such as the Deep Research feature of OpenAI \cite{openaidr} to emerging open-source agents such as Tongyi-DeepResearch \cite{tongyidrs} and MiroThinker \cite{mirothinker}.

However, existing open-source web agents remain highly inefficient. For example, systems such as MiroThinker often require several hundred reasoning rounds to answer a single question. This inefficiency arises from training paradigms that incentivize brute-force strategies, in which task success is pursued by indiscriminately increasing tool usage. The resulting volume of unnecessary tool calls imposes substantial computational and time overhead: executing tools at scale requires considerable infrastructure resources and increases operational costs.

Our analysis attributes this overall inefficiency to two distinct failure modes that arise at different stages of the agent workflow: \textit{blind tool dependency} and \textit{performative reasoning}. First, \textbf{blind tool dependency} reflects a failure of initiation and functions as a form of indiscriminate cognitive offloading \cite{risko2016cognitive}. Agents not only overuse tools but also invoke external tools for simple queries that can be resolved utilizing internal knowledge, thereby consuming API calls redundantly and introducing unnecessary latency rather than directly producing an answer. Second, \textbf{performative reasoning} emerges in complex tasks that legitimately require external tools and represents a critical failure of execution efficiency. Rather than identifying a direct solution, the reasoning process of the agent becomes saturated with redundant loops, repetitive verifications of previously established facts, and unproductive exploratory branches. Consequently, models generate protracted trajectories that mimic rigor yet add no substantive information to the final answer. Collectively, these issues indicate a fundamental deficiency in the ability of the model to extract the Minimal Necessary Path \cite{wang2026webclipperefficientevolutionweb}.

We find that these defects are exacerbated by the prevailing \textbf{Acc-only Rejection Sampling} paradigm, where any correct trajectory, regardless of its redundancy, is treated as a positive sample for SFT. Naively training on such data causes models to internalize bloated patterns (see our experiments), as standard rejection sampling provides no mechanism to filter out these hollow trajectories. Furthermore, existing Reinforcement Learning (RL) strategies for web agents\cite{jin2025Search-r1, li2025torl} primarily incentivize end-to-end success rates. This singular focus triggers an ``efficiency collapse'' during RL training: as iterations progress, the model tends to scale up search rounds and context lengths, relying on over-exploration to maximize accuracy.

To address this efficiency trap, we propose \textbf{SlimSearcher}, a framework designed for long-horizon deep research scenarios that systematically integrates efficiency optimization into both SFT and RL to push the Pareto frontier between accuracy and computational cost. Our core philosophy is that the system must not only guide the model to the correct answer but also progressively converge toward the  Minimal Necessary
Path. Specifically, our contributions are summarized as follows:

\begin{itemize} 

\item \textbf{Efficiency-Aware SFT:} Instead of filtering trajectories by correctness alone, we introduce a joint efficiency evaluation to construct a refined training dataset. This ensures the model learns efficient search behaviors rather than redundant patterns.

\item \textbf{Adaptive Efficiency Anchoring (AEA) for Reward Shaping:} To circumvent the brevity bias inherent in fixed penalties, we design a dynamic reward structure that explicitly anchors optimization to the empirical  Minimal Necessary
Path discovered within a {sampled trajectory group}. This approach provides a task-adaptive gradient signal, incentivizing convergence toward minimal trajectory complexity without compromising task accuracy.

\item \textbf{Efficiency-Accuracy Pareto Improvement:} Extensive experiments on  long-horizon benchmarks, including GAIA, BrowseComp, and XBench-DeepSearch, demonstrate SlimSearcher's strong efficiency gains. Compared to baselines, SlimSearcher reduces average tool-call rounds by \textbf{17\%--58\%} while improving task accuracy. 
\end{itemize}

\begin{figure*}
    \centering
    \includegraphics[width=\linewidth]{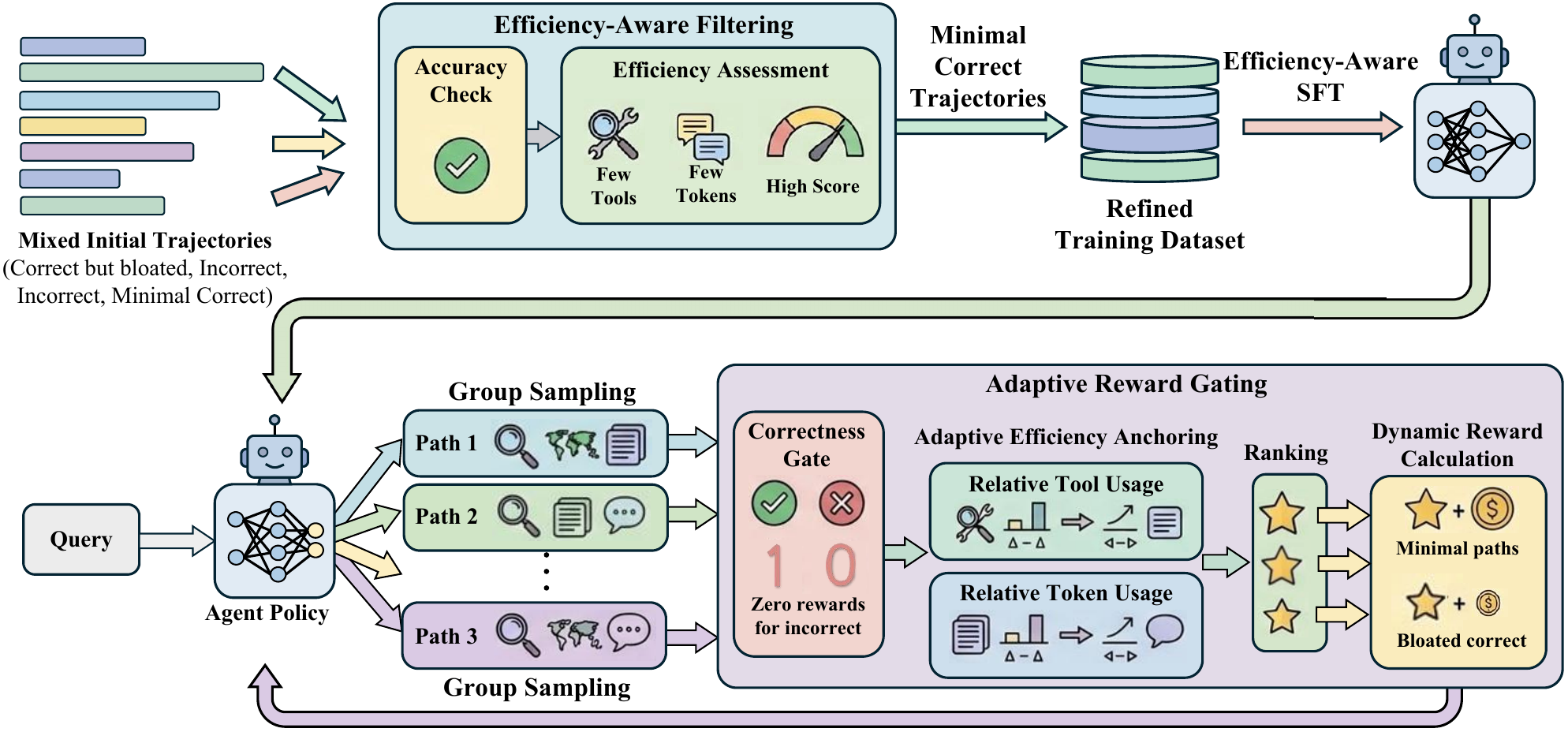}
\caption{SlimSearcher: Multi-dimensional SFT filtering and adaptive reward gating enable web agents to learn efficiency-aware and correct search behaviors.}
    \label{fig:main_graph}
\end{figure*}

\section{Related Work}

\subsection{Paradigm of Web Agents}
Current methodologies for autonomous web agents can be broadly categorized into two streams. The first stream is \textit{Prompting and Engineering Frameworks}, which elicit agentic behaviors from frozen LLMs through collaborative architectures. This includes classical prompting strategies like SelfDC~\cite{wang2025self}, as well as multi-agent frameworks such as AutoGen and GPT-Researcher. While flexible, these systems rely heavily on the inherent capabilities of the base model and often incur high token costs due to verbose context management.
The second stream is \textit{Training-Centric Agent Evolution}, which internalizes search and reasoning capabilities directly into model parameters via Supervised Fine-Tuning (SFT)~\cite{webleaper} or Reinforcement Learning (RL). Early works focused on synthesizing trajectories for SFT, such as WebSailor \cite{li2025websailor} and WebShaper \cite{tao2507webshaper}. More recent approaches leverage RL to enhance performance, as seen in Search-R1 \cite{jin2025Search-r1}. However, most training-based methods prioritize success rates as the sole metric, often incentivizing agents to adopt ``brute-force'' strategies that exhaust computational resources to force a correct answer, largely ignoring inference efficiency and computational cost. Existing methods reduce internal reasoning length through prompt engineering \cite{han2025token,xu2025chain} or by penalizing long Chain-of-Thought (CoT) paths during RL \cite{ma2025cot,munkhbat2025self,aggarwal2025l1}. However, optimizing web agents requires more than reducing text tokens; it requires systematically pruning redundant external actions---such as cyclic searches or unproductive browsing---which are the primary bottleneck in web agents.

\subsection{Efficiency in Tool-Integrated Reasoning}

As agents tackle increasingly complex, long-horizon tasks, execution efficiency has emerged as a critical dimension. In the realm of SFT, \textbf{WebLeaper}~\cite{webleaper} argues that sparse entity tasks in training data limit the agent's ability to learn efficient search behaviors, proposing the synthesis of high-density information-seeking tasks to address this gap. Unlike WebLeaper, which focuses primarily on data synthesis, \textbf{SlimSearcher} fundamentally advances this paradigm by unifying efficiency optimization across the entire training pipeline . To achieve this, SlimSearcher introduces \textbf{Adaptive Efficiency Anchoring (AEA)}. Instead of applying static cost constraints that often induce brevity bias, AEA dynamically calibrates the reward landscape by anchoring it to the empirical minimal necessary path discovered within a sampled trajectory group. Furthermore, we enforce a strict correctness gate to inherently prevent reward hacking. This synergistic design ensures that SlimSearcher autonomously shifts the Pareto frontier, rigorously preserving task accuracy while systematically eradicating computational pleonasm.

\section{Methodology}
\label{sec:methodology}

In this section, we present the \textbf{SlimSearcher} framework. Unlike previous methods that rely on heuristic prompt engineering, SlimSearcher unifies the optimization of accuracy and efficiency into a single \textbf{Multi-Stage Gating} mechanism. This mechanism serves a dual purpose: it acts as a \textit{Pareto Filter} during the SFT phase to construct high-signal demonstration data via Rejection Sampling, and as a \textit{Cascading Reward Function} during the RL phase to guide policy evolution.

\subsection{The Multi-Stage Gating Mechanism}
\label{sec:gating_mechanism}

The core of SlimSearcher is a hierarchical valuation system. Its primary objective is to purge ``unproductive trajectories''---paths that are technically correct but marred by \textit{blind tool dependency} or \textit{performative reasoning}. Drawing inspiration from the convergence mechanisms in ant colony optimization, our reward shaping is designed to iteratively approximate the {Minimal Necessary Path} . By progressively driving the policy toward the shortest viable route over successive explorations, we ensure agents operate on the efficient Pareto frontier. Formally, given a question $q$ and a generated trajectory $\tau$, the final reward $R_{final}$ is computed via a multiplicative cascading logic:

\begin{equation} \label{final_reward}
    R_{final}(\tau) = r_{correct}(\tau) \cdot r_{tool}(\tau) \cdot r_{len}(\tau)
\end{equation}

\subsubsection{Gate 1: The Correctness Gate}
The first gate imposes a strict binary constraint. We define $r_{correct}$ as:
\begin{equation}
    r_{correct} = \mathbb{I}(\text{Answer}(\tau) \approx y_{gt})
\end{equation}
If the trajectory fails to recover the ground truth $y_{gt}$, $r_{correct} = 0$, forcing the entire chain $R_{final}$ to zero. This ensures that the model never prioritizes brevity over accuracy, strictly preventing the ``reward hacking'' phenomenon where models produce short but hallucinatory answers.

\subsubsection{Gate 2: Adaptive Anchoring for Tool Efficiency}
For trajectories that pass the Correctness Gate, we introduce an adaptive scaling mechanism for tool usage. Instead of applying a fixed linear penalty to tool calls---which often provides weak gradient signals \cite{wu2024dler} and induces \textit{brevity bias} \cite{zhang2024agentic}---we dynamically anchor the reward to the most efficient trajectory discovered within the current exploration batch. Analogous to ant colony optimization, where the shortest discovered route receives the strongest reinforcement, we use the empirical minimum cost to shape the reward landscape.

Specifically, a trajectory $\tau_i$ consists of a sequence of reasoning steps and actions. We denote the set of external tool invocations within $\tau_i$ as $\mathcal{A}(\tau_i)$. The total tool cost $C(\tau_i)$ is formulated as the weighted sum of all executed tool calls:
\begin{equation}
    C(\tau_i) = \sum_{a \in \mathcal{A}(\tau_i)} w_{type}(a)
\end{equation}
where $w_{type}(a)$ represents the predefined penalty weight corresponding to the specific tool type of action $a$. For simplicity, we set $w_{type}(a) = 1$ uniformly for all tool types in our experiments.

Given a candidate set of sampled trajectories for a specific query, we identify the lowest cost among the successful paths as the empirical minimum, $C_{\min}$. To measure the inefficiency relative to this minimal-cost path, we formulate the relative deviation $\delta_i$ and the associated score $S_i$ as follows:

\begin{equation}
    \delta_i = \frac{C(\tau_i) - C_{min}}{C_{min} + \epsilon}, \quad S_i = -\delta_i
\end{equation}
where $\epsilon$ is a small constant for numerical stability (e.g., $10^{-8}$). Rather than using $S_i$ directly, we map this relative score into a bounded reward space using an exponential transformation:
\begin{equation}
    r_{tool}(\tau_i) = 2 \cdot \frac{\exp(S_i)}{\exp(S_i) + \exp(S_{opt})}
\end{equation}
where $S_{opt}=0$ represents the optimal behavior in the candidate set (i.e., when $C(\tau_i) = C_{min}$). 

This design choice is deliberate: much like pheromone trails that dissipate rapidly over suboptimal routes, the non-linear mapping ensures that trajectories diverging from the empirical optimal path suffer an exponentially decaying penalty. By converting absolute tool costs into a strictly bounded multiplier ($r_{tool} \in (0, 1]$), we provide a sharper and more stable efficiency signal prior to the advantage estimation computed later during the RL phase.

\subsubsection{Gate 3: Adaptive Anchoring  for Token Efficiency}
While Gate 2 optimizes for external tool costs, it does not prevent the model from generating excessive internal tokens to compensate for fewer observations. To address this \textit{performative reasoning}, we introduce the \textbf{Adaptive Efficiency Anchoring} mechanism to impose a dynamic length constraint relative to the candidate set's most concise successful solution.

Let $\mathcal{G}_{correct}$ denote the subset of concurrent trajectories that passed Gate 1. We define the baseline minimal length $L_{min}$:
\begin{equation}
    L_{min} = \min_{\tau \in \mathcal{G}_{correct}} L(\tau)
\end{equation}
For any trajectory $\tau_i$, we calculate the relative length deviation $\rho_i$ and the score $S_{len, i}$, quantifying the redundant token consumption:
\begin{equation}
    \rho_i = \frac{L(\tau_i) - L_{min}}{L_{min} + \epsilon}, \quad S_{len, i} = -\rho_i
\end{equation}
Applying the same non-linear transformation used in Gate 2, we map this score into a bounded multiplier:
\begin{equation}
    r_{len}(\tau_i) = 2 \cdot \frac{\exp(S_{len, i})}{\exp(S_{len, i}) + \exp(S_{opt})}
\end{equation}
where $S_{opt}=0$. This parallel formulation guarantees that overly verbose trajectories face a strict, non-linear penalization proportional to their excess length. By continuously reinforcing the most compact generation within the competitive cohort, the policy progressively suppresses performative bloated patterns, driving the agent toward truly parsimonious reasoning.

\subsection{Efficiency-Aware SFT via Reward-Guided Rejection Sampling}
\label{sec:stage1_sft}

We employ the base model as our backbone. While this model possesses robust reasoning capabilities, its raw outputs often suffer from \textit{verbosity bias}---generating trajectories that are factually correct yet computationally inefficient. To align the model with our efficiency objectives, we construct a high-signal dataset, $\mathcal{D}_{sft}$, through a rigorous \textit{Pareto-Efficient Filtration} pipeline, acting as a reward-guided Rejection Fine-Tuning process.

Specifically, we compile a comprehensive seed corpus comprising 13,863 high-quality trajectories sourced from a diverse array of information-seeking datasets (\ref{sec:SFT_stage}), alongside synthetically generated instances. To ensure the SFT phase targets non-trivial reasoning, we evaluate query difficulty by executing each prompt four times in the base environment. We retain only queries $q$ exhibiting a pass rate $\mathrm{PR}(q)$ within the interval $0 < \mathrm{PR}(q) < 1$, effectively excluding trivial queries requiring no deep research and impossible tasks that might introduce noise. 
For each filtered query $q$, we sample a candidate set of $K$ trajectories, denoted as $\mathcal{T}_q = \{\tau_1, \tau_2, \dots, \tau_K\}$, using the base model as the generator, and apply our Multi-Stage Gating mechanism. First, we filter out all trajectories that fail to recover the ground truth by applying the correctness indicator, yielding a valid subset $\mathcal{T}_q^{valid}$:
\begin{equation}
    \mathcal{T}_q^{valid} = \{\tau \in \mathcal{T}_q \mid r_{correct}(\tau) = 1\}
\end{equation}
Subsequently, we evaluate the remaining valid candidates based on their joint efficiency score, defined as the product of the tool-call efficiency $r_{tool}$ and the token efficiency $r_{len}$. We distill the Minimal Necessary Path (MNP), denoted as $\tau^*$, by selecting the trajectory that maximizes this joint objective:
\begin{equation}
    \tau^* = \underset{\tau \in \mathcal{T}_q^{valid}}{\arg\max} \big( r_{tool}(\tau) \times r_{len}(\tau) \big)
\end{equation}
This rigorous filtration effectively purges redundant or poisoned reasoning loops. The final collection of these optimal trajectories $\tau^*$ forms our refined dataset $\mathcal{D}_{sft}$, which is explicitly optimized to teach parsimonious and cost-aware search behaviors right from the initialization stage.

\begin{table*}[t]
\centering
\definecolor{softblue}{RGB}{235, 240, 245}

\renewcommand{\arraystretch}{1.2} 
\setlength{\tabcolsep}{4pt} 

\resizebox{\textwidth}{!}{
\begin{tabular}{l ccc ccc ccc ccc}
\toprule
\multirow{2}{*}{\textbf{Method}} & \multicolumn{3}{c}{\textbf{XBench-DeepSearch}} & \multicolumn{3}{c}{\textbf{Browsecomp}} & \multicolumn{3}{c}{\textbf{GAIA}} & \multicolumn{3}{c}{\textbf{HLE}} \\
\cmidrule(lr){2-4} \cmidrule(lr){5-7} \cmidrule(lr){8-10} \cmidrule(lr){11-13}
& \cellcolor{softblue}\textbf{Acc} $\uparrow$ & \textbf{Rounds} $\downarrow$ & \textbf{Token} $\downarrow$ 
& \cellcolor{softblue}\textbf{Acc} $\uparrow$ & \textbf{Rounds} $\downarrow$ & \textbf{Token} $\downarrow$ 
& \cellcolor{softblue}\textbf{Acc} $\uparrow$ & \textbf{Rounds} $\downarrow$ & \textbf{Token} $\downarrow$ 
& \cellcolor{softblue}\textbf{Acc} $\uparrow$ & \textbf{Rounds} $\downarrow$ & \textbf{Token} $\downarrow$ \\
\midrule

\multicolumn{13}{l}{\textit{\textbf{Proprietary Systems}}} \\
OpenAI o3            
& \cellcolor{softblue}0.670 & - & - 
& \cellcolor{softblue}\underline{0.497} & - & - 
& \cellcolor{softblue}- & - & - 
& \cellcolor{softblue}0.249 & - & - \\
OpenAI DeepResearch 
& \cellcolor{softblue}- & - & - 
& \cellcolor{softblue}\textbf{0.515} & - & - 
& \cellcolor{softblue}0.674 & - & - 
& \cellcolor{softblue}0.266 & - & - \\
Claude-4-Sonnet     
& \cellcolor{softblue}0.646 & - & - 
& \cellcolor{softblue}0.122 & - & - 
& \cellcolor{softblue}0.683 & - & - 
& \cellcolor{softblue}0.203 & - & - \\
\midrule

\multicolumn{13}{l}{\textit{\textbf{Open-weights Agents}}} \\

Kimi-K2-Instruct-0905 
& \cellcolor{softblue}0.540 & 5.98 & 1316 
& \cellcolor{softblue}0.094 & 16.65 & 3426 
& \cellcolor{softblue}0.469 & \underline{6.45} & 1281 
& \cellcolor{softblue}0.146 & \textbf{5.17} & \textbf{2349} \\
Qwen3-235B-A22B-Instruct 
& \cellcolor{softblue}0.490 & 8.84 & \underline{938}  
& \cellcolor{softblue}0.046 & \underline{13.70} & 1837 
& \cellcolor{softblue}0.456 & 7.14 & 1128 
& \cellcolor{softblue}0.199 & \underline{7.45} & 2960 \\
DeepSeek-V3
& \cellcolor{softblue}0.530 & \textbf{2.23} & \textbf{422}
& \cellcolor{softblue}0.038 & \textbf{4.78} & \textbf{1317}
& \cellcolor{softblue}0.373 & \textbf{3.20} & \underline{994}
& \cellcolor{softblue}0.060 & 13.00 & 2713 \\

WebExplorer         
& \cellcolor{softblue}0.517 & 9.05 & 2246 
& \cellcolor{softblue}0.137 & 29.43 & 6289 
& \cellcolor{softblue}0.372 & 12.88 & 3560 
& \cellcolor{softblue}0.116 & 15.52 & 6579 \\

WebLeaper
& \cellcolor{softblue}\underline{0.780} & 14.06 & 4466.50
& \cellcolor{softblue}0.388 & 60.07 & 10894
& \cellcolor{softblue}\underline{0.702} & 16.65 & 5045
& \cellcolor{softblue}0.298 & 24.86 & 11924 \\

\midrule

Qwen3-30B-A3B-Instruct-2507$^*$ 
& \cellcolor{softblue}0.550 & 15.04 & 2395 
& \cellcolor{softblue}0.029 & 25.80 & \underline{1683} 
& \cellcolor{softblue}0.339 & 11.63 & \textbf{986}  
& \cellcolor{softblue}0.009 & 25.77 & \underline{2570} \\
\midrule
\multicolumn{13}{l}{\textit{\textbf{Baseline: Qwen3-30B-A3B-Instruct-2507}}} \\
SlimSearcher (SFT)                 
& \cellcolor{softblue}0.740 & 13.44 & 5515 
& \cellcolor{softblue}0.309 & 65.53 & 12185 
& \cellcolor{softblue}0.665 & 24.46 & 7299 
& \cellcolor{softblue}0.259 & 27.86 & 13169 \\

\rowcolor{gray!10} 
SlimSearcher (SFT+RL)   
& \cellcolor{softblue}0.770 & 10.87 & 5092 
& \cellcolor{softblue}0.328 & 47.22 & 10883  
& \cellcolor{softblue}0.699 & 20.13 & 7237 
& \cellcolor{softblue}0.278 & 19.51 & 11497 \\

\midrule

Tongyi-DeepResearch 
& \cellcolor{softblue}0.713 & 14.26 & 6918 
& \cellcolor{softblue}0.410 & 63.70 & 12014 
& \cellcolor{softblue}0.682 & 20.56 & 7378 
& \cellcolor{softblue}0.358 & 23.92 & 13664 \\
\midrule
\multicolumn{13}{l}{\textit{\textbf{Baseline: Tongyi-DeepResearch}}} \\
Prompt Control      
& \cellcolor{softblue}0.676 & 12.50 & 6321 
& \cellcolor{softblue}0.373 & 62.80 & 12222 
& \cellcolor{softblue}0.663 & 18.70 & 6752 
& \cellcolor{softblue}0.349 & 23.91 & 14107 \\
SlimSearcher (SFT)      
& \cellcolor{softblue}0.770 & 10.22 & 5155 
& \cellcolor{softblue}0.428 & 61.28 & 11522 
& \cellcolor{softblue}0.660 & 21.79 & 6412 
& \cellcolor{softblue}\underline{0.366} & 27.16 & 12936 \\
\rowcolor{gray!10} 
SlimSearcher (SFT+RL)   
& \cellcolor{softblue}\textbf{0.790} & \underline{5.92} & 4888 
& \cellcolor{softblue}0.447 & 47.63 & 11093 
& \cellcolor{softblue}\textbf{0.709} & 10.61 & 4915 
& \cellcolor{softblue}\textbf{0.376} & 19.82 & 11418 \\

\bottomrule
\end{tabular}}
\caption{\textbf{Main results across diverse web agent benchmarks.} Evaluation metrics include task accuracy ($\text{Acc} \uparrow$), interaction rounds ($\text{Rounds} \downarrow$), and total token consumption ($\text{Token} \downarrow$). \textbf{Bold} and \underline{underline} indicate the best and second-best results, respectively, across all evaluated models for each metric. Our proposed method (\textbf{SlimSearcher}) demonstrates a Pareto improvement in both task completion and resource economy across all benchmarks. $^*$ denotes results evaluated in our unified environment.}
\label{tab:results_all_benchmarks}
\vspace{-3mm}
\end{table*}

\subsection{Policy Optimization with Multi-Stage Gating Reward}
\label{sec:stage2_rl}

We further employ GRPO\cite{DeepSeek-R1}-based reinforcement learning with the Multi-Stage Gating reward to guide the model in exploring the Pareto-optimal frontier between efficiency and accuracy.

For each input question $q \sim \mathcal{D}$, we sample a group of $G$ trajectories $\{\tau_1, \tau_2, \dots, \tau_G\}$ from the old policy $\pi_{old}$. The raw reward for each trajectory is computed using our cascading mechanism: $R_i = R_{final}(\tau_i)$, which is the same as Equation (\ref{final_reward}). 

We then estimate the group-relative advantage $\hat{A}_i$ for each trajectory by standardizing its reward against the group's performance:
\begin{equation}
    \hat{A}_i = \frac{R_i - \text{mean}(\{R_1, \dots, R_G\})}{\text{std}(\{R_1, \dots, R_G\}) + \epsilon_s}
\end{equation}
where $\epsilon_s$ is a small constant for numerical stability. This trajectory-level advantage $\hat{A}_i$ is applied to every generated token at timestep $t$ within the trajectory $\tau_i$. The policy model $\pi_\theta$ is updated by maximizing the clipped surrogate objective function:
\begin{equation}
\small
\begin{split}
    &\mathcal{J}_{GRPO}(\theta) = \mathbb{E}_{q \sim \mathcal{D}, \{\tau_i\}_{i=1}^G \sim \pi_{old}} \Bigg[ \frac{1}{G} \sum_{i=1}^G \frac{1}{|\tau_i|} \sum_{t=1}^{|\tau_i|} \\
    & \quad \min \Big( p_{i,t} \hat{A}_i, \text{clip}(p_{i,t}, 1-\epsilon_c, 1+\epsilon_c) \hat{A}_i \Big) \Bigg] 
\end{split}
\end{equation}
where $p_{i,t} = \frac{\pi_\theta(y_{i,t} | q, y_{i,<t})}{\pi_{old}(y_{i,t} | q, y_{i,<t})}$ represents the importance sampling ratio of generating token $y_{i,t}$ given the query $q$ and preceding tokens $y_{i,<t}$ at timestep $t$, $\epsilon_c$ is the clipping bound to ensure stable updates.


\section{Experiments}

\subsection{Experimental Settings}

\textbf{ Datasets.}
We conduct experiments on four widely adopted web agent benchmarks: XBench-DeepSearch \cite{xbench}, BrowseComp \cite{bc_en}, GAIA \cite{mialon2023gaia}, and HLE \cite{hle}, following the same set as \cite{wang2026webclipperefficientevolutionweb}. More details can be found in Appendix \ref{sec:dataset}.

\textbf{Evaluation Metrics.} To comprehensively assess the trade-off between task success and operational expenditure, we employ a multi-dimensional framework. We measure \textbf{Accuracy (Acc)} as the primary success rate, representing the percentage of queries where the final response aligns with the ground truth. To quantify efficiency, we track \textbf{Tool-Call Rounds}, the average number of external tool invocations per query, which directly reflects the model's ability to mitigate \textit{blind tool dependency}. Additionally, we monitor \textbf{Token Efficiency} via the average consumption of reasoning tokens per query. Note that this metric is strictly limited to model-generated tokens . This metric serves as a key indicator of \textit{performative reasoning} and overall computational overhead.

\textbf{Baselines.}
Our comparison includes both closed-source and open-source agents. Closed-source systems include OpenAI o3~\cite{o3}, OpenAI DeepResearch~\cite{openaidr}, and Claude-4-Sonnet~\cite{claude4}, with test results cited from their official reports. The open-source agents include Kimi-K2-Instruct-0905~\cite{kimik2}, Qwen3-235B-A22B-Instruct-2507~\cite{yang2025qwen3technicalreport}, DeepSeek-V3~\cite{deepseekai2025deepseekv3technicalreport}, WebExplorer~\cite{webexplorer}, and WebLeaper~\cite{webleaper}. To verify the effectiveness of our approach, we design the following baseline: \textit{Prompt Control}, where we add instructions to the agent's system prompt, explicitly asking it to avoid irrelevant information and repetitive validation while controlling the number of tool calls.


\paragraph{Implementation.} 
Our backbone policies build on \textit{Tongyi-DeepResearch-30B} and \textit{Qwen3-30B-A3B-Instruct-2507}. We distribute computation across 64 NVIDIA H800 GPUs for SFT and 64 H800 GPUs for RL to meet the high rollout throughput required by GRPO. For environment interaction, we use the Serper API for search and Jina Reader for URL parsing. To improve statistical reliability, we run each evaluation three times and report the average Pass@1 and efficiency metrics. Detailed hyperparameters are provided in Appendix~\ref{sec:Implementation}.
\begin{figure*}[t]
    \centering
    \includegraphics[width=\textwidth]{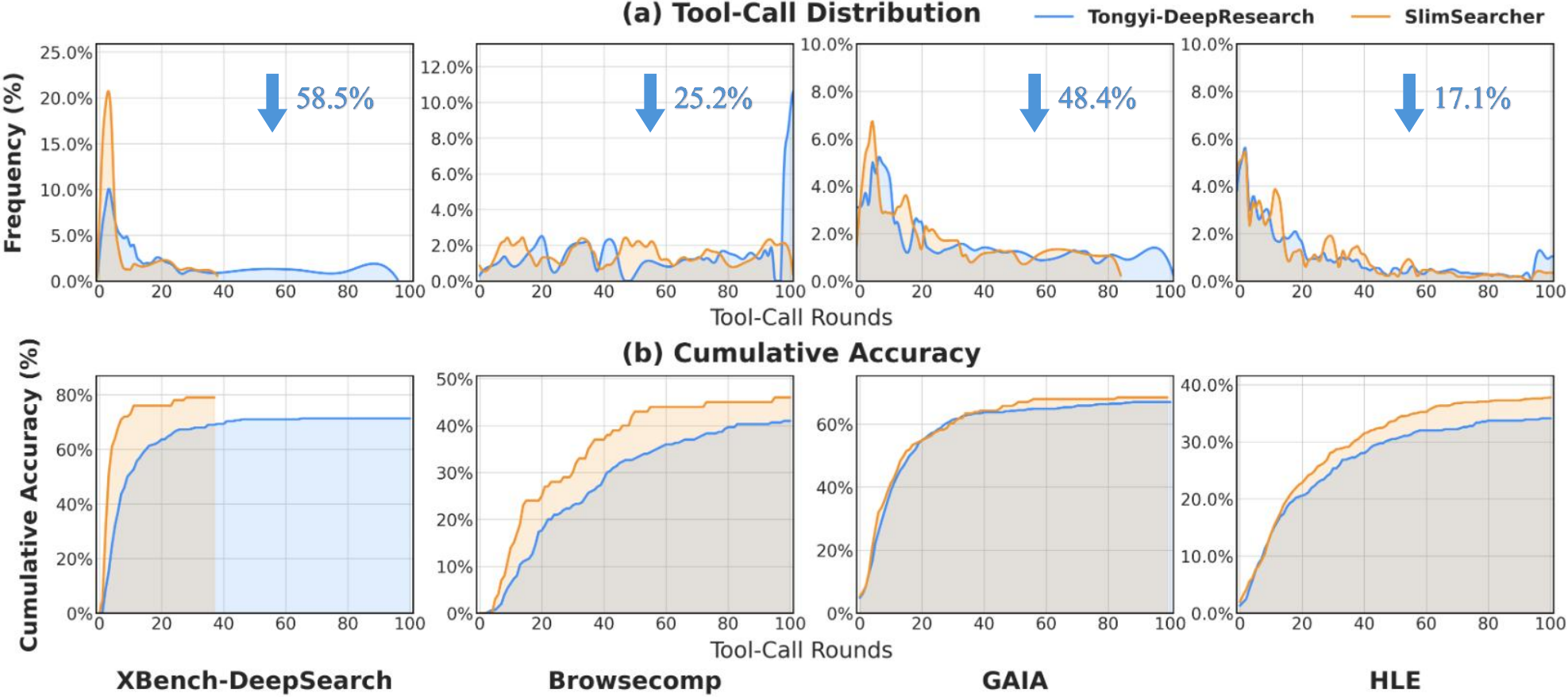}
    \caption{Comparison of tool-call distribution and cumulative accuracy.}
    \label{fig:tool_distribution}
\end{figure*}

\subsection{Main Results}

Table \ref{tab:results_all_benchmarks} reports the performance of SlimSearcher and baseline models on benchmarks. The results show that SlimSearcher improves efficiency while maintaining accuracy across different backbones.

For the \textbf{Tongyi-DeepResearch} backbone, SlimSearcher (SFT+RL) reduces computational overhead while maintaining or improving accuracy. On GAIA, tool-call rounds decrease by 48.4\% (from 20.56 to 10.61) and token usage decreases by 33.4\%, while accuracy increases from 0.682 to 0.709. On Browsecomp, tool-call rounds decrease from 63.70 to 47.63 and token usage decreases from 12014 to 11093, while accuracy increases from 0.410 to 0.447.

For the \textbf{Qwen3-30B-A3B-Instruct} backbone, the base model shows low accuracy. Fine-tuning on our high-quality, Pareto-filtered dataset enables deeper research behavior and improves accuracy . This stage increases token usage to 7299 because the model learns to execute more complex searches, but the subsequent RL stage shortens the trajectory. On HLE, the RL stage reduces tool-call rounds from 27.86 to 19.51 and increases accuracy from 0.259 to 0.278.
In contrast,  Prompt Control fails to consistently improve efficiency and actually reduces accuracy. Similarly, baselines like DeepSeek-V3 consume minimal tokens but fail to achieve competitive accuracy. Overall, these results demonstrate that coupling Pareto-filtered SFT data with Adaptive Reward Gating enables SlimSearcher to prune redundant tool calls without sacrificing task accuracy.

\definecolor{softblue}{RGB}{235, 240, 245}

\renewcommand{\arraystretch}{1.2}
\setlength{\tabcolsep}{4pt}

\begin{table*}[t]
\centering
\resizebox{\textwidth}{!}{
\begin{tabular}{l ccc ccc ccc ccc}
\toprule
\multirow{2}{*}{\textbf{Method}} & \multicolumn{3}{c}{\textbf{XBench-DeepResearch}} & \multicolumn{3}{c}{\textbf{Browsecomp}} & \multicolumn{3}{c}{\textbf{GAIA}} & \multicolumn{3}{c}{\textbf{HLE}} \\
\cmidrule(lr){2-4} \cmidrule(lr){5-7} \cmidrule(lr){8-10} \cmidrule(lr){11-13}
& \cellcolor{softblue}\textbf{Acc} $\uparrow$ & \textbf{Rounds} $\downarrow$ & \textbf{Token} $\downarrow$ 
& \cellcolor{softblue}\textbf{Acc} $\uparrow$ & \textbf{Rounds} $\downarrow$ & \textbf{Token} $\downarrow$ 
& \cellcolor{softblue}\textbf{Acc} $\uparrow$ & \textbf{Rounds} $\downarrow$ & \textbf{Token} $\downarrow$ 
& \cellcolor{softblue}\textbf{Acc} $\uparrow$ & \textbf{Rounds} $\downarrow$ & \textbf{Token} $\downarrow$ \\
\midrule

\multicolumn{13}{l}{\textit{\textbf{Base Model}}} \\
Qwen3-30B-A3B-Instruct 
& \cellcolor{softblue}0.550 & 15.04 & 2395
& \cellcolor{softblue}0.029 & 25.80 & 1683
& \cellcolor{softblue}0.339 & 11.63 & 986 
& \cellcolor{softblue}0.009 & 25.77 & 2570 \\

\midrule

\multicolumn{13}{l}{\textit{\textbf{Stage 1: SFT Phase}}} \\
SFT (Standard Rejection Sampling)    
& \cellcolor{softblue}0.730 & 14.08 & {5338} 
& \cellcolor{softblue}0.287 & 66.08 & {12065} 
& \cellcolor{softblue}0.641 & 25.90 & 7478
& \cellcolor{softblue}0.262 & 29.35 & {12724} \\

SFT (Reward-Guided Rejection Sampling)     
& \cellcolor{softblue}0.740 & {13.44} & 5515 
& \cellcolor{softblue}0.309 & {65.53} & 12185 
& \cellcolor{softblue}{0.665} & {24.46} & {7299}
& \cellcolor{softblue}0.259 & {27.86} & 13169 \\

\midrule

\multicolumn{13}{l}{\textit{\textbf{Stage 2: RL Phase}}} \\
\quad \textit{w/o Correctness Gate} 
& \cellcolor{softblue}0.440 & 1.42  & 3022      
& \cellcolor{softblue}0.007 & 0.20  & 7395
& \cellcolor{softblue}0.136 & 0.07  & 4239 
& \cellcolor{softblue}0.124 & 1.17  & 4898 \\     



\quad \textit{w/o Adaptive Efficiency Anchoring}
 & \cellcolor{softblue}{0.780}  & 15.54   &  5671
 & \cellcolor{softblue}{0.339} &  74.10 &  13644
& \cellcolor{softblue}0.641 &  25.42 & 7409
& \cellcolor{softblue}{0.283} &  31.05   &  13781
\\

\rowcolor{gray!10}
\textbf{SlimSearcher (Full)}
& \cellcolor{softblue}{0.770} & {10.87} & {5092}
& \cellcolor{softblue}{0.328} & {47.22} & {10883} 
& \cellcolor{softblue}{0.699} & {20.13} & {7237} 
& \cellcolor{softblue}{0.278} & {19.51} & {11497} \\

\bottomrule
\end{tabular}} 
\caption{Performance comparison of different training strategies across four long-horizon web research benchmarks. }
\label{tab:data_ablation}
\vspace{-3mm}
\end{table*}

\subsection{Detailed Performance Analysis Across Benchmarks}
\label{sec:detailed_performance}

To further assess the robustness and efficiency of SlimSearcher, we conduct a fine-grained analysis of the distribution of search rounds and cumulative accuracy  across four representative benchmarks. As shown in Figure \ref{fig:tool_distribution}, SlimSearcher consistently shifts the Pareto frontier, achieving higher accuracy with a substantially more compact action space.

\paragraph{GAIA: Strict Execution Discipline in Complex Reasoning} 
The GAIA dataset requires complex, multi-step reasoning. Notably, in such reasoning-intensive scenarios, many queries can actually be resolved utilizing the model's internal knowledge without relying on external tools. However, baseline models struggle significantly here, exhibiting "blind tool dependency" and scattered, inefficient tool usage  when internal capabilities would suffice. In contrast, SlimSearcher enforces strict reasoning discipline. By mitigating this indiscriminate cognitive offloading , it resolves most queries on GAIA within the 0--20 round bucket, driving a steep accuracy curve that approaches 70\%. This demonstrates the framework's ability to effectively leverage its inherent capabilities, eliminate invalid trial-and-error steps, and strategically target core information through external tools only when necessary.

\paragraph{XBench, Browsecomp, and HLE: Eliminating the Efficiency Trap in Exploration}
Beyond deep reasoning, web agents also face severe challenges in open-domain search, multi-step page navigation, and extreme long-horizon tasks, which SlimSearcher consistently overcomes across the remaining benchmarks. The XBench dataset involves open-ended web search and information extraction. On this dataset, the baseline model's tool-call distribution exhibits a long, inefficient tail, indicating that the agent frequently gets caught in unproductive verification loops and cyclic behavior. SlimSearcher markedly compresses this distribution, sharply concentrating almost all tool calls within the 0--20 round interval with a massive peak under 10 rounds. The cumulative accuracy curve shows that SlimSearcher reaches peak performance early, which confirms that the proposed framework removes performative reasoning loops while improving task success. Similarly, Browsecomp requires multi-step web navigation, where agents can easily get lost in irrelevant exploratory branches. The distribution for both the Base models shows a long, inefficient tail, with a massive spike of failures hitting the $>100$ round budget. SlimSearcher suppresses this long tail. It densely concentrates its tool calls in the 20--60 round interval, and its cumulative accuracy at 40 rounds already eclipses the final accuracy of the baseline at 100 rounds. Finally, HLE requires complex, multi-step reasoning where baseline models suffer complete failure by hitting the $>100$ round limit with flat accuracy. On the inherently difficult HLE dataset, SlimSearcher bounds its exploration within the first 40 rounds, maintaining a steady accuracy growth that consistently dominates the baselines.

\subsection{Ablation Study}
\label{sec:ablation}

\paragraph{Effectiveness of Reward-Guided Rejection Sampling.}
Unlike \textit{Standard Rejection Sampling}, which indiscriminately accepts any correct trajectory and inadvertently internalizes bloated reasoning patterns, our \textit{Reward-Guided} approach enforces a rigorous joint efficiency evaluation. As demonstrated in Table \ref{tab:data_ablation}, this mechanism yields consistent Pareto improvements across benchmarks. For instance, on GAIA, it not only elevates task accuracy ($0.641 \rightarrow 0.665$) but also systematically compresses the trajectory footprint, reducing average tool-call rounds ($25.90 \rightarrow 24.46$) and token consumption ($7478 \rightarrow 7299$).

\paragraph{Necessity of the Correctness Gate.} 
Removing the strict correctness constraint during RL induces a catastrophic accuracy collapse accompanied by an anomalous plunge in tool usage. As shown in Table \ref{tab:data_ablation}, without this gate, the agent's performance on GAIA drops significantly in accuracy (0.136) while registering near-zero tool exploration (0.07 rounds). This indicates severe reward hacking: in the absence of an absolute success prerequisite, the policy optimizes solely for the efficiency reward by generating immediate, vacuous responses, bypassing the reasoning process entirely.

\paragraph{Impact of Adaptive Efficiency Anchoring.}
The ablation of our dynamic scaling mechanism (\textit{w/o Adaptive Efficiency Anchoring}) severely disrupts the delicate equilibrium between task efficacy and computational cost. Without the empirical minimal path serving as a dynamic anchor, the model defaults to brute-force exploration. On HLE, this unrestricted search space inflates tool-call rounds significantly ($19.51 \rightarrow 31.05$) to squeeze out a marginal accuracy gain , incurring  computational overhead. Furthermore, the absence of this joint anchoring degrades internal reasoning discipline; on reasoning-intensive benchmarks like GAIA, accuracy actually deteriorates ($0.699 \rightarrow 0.641$) despite increased tool usage ($20.13 \rightarrow 25.42$), underscoring the necessity of AEA in suppressing performative reasoning.


\section{Conclusion}


In this paper, we address the efficiency trap in web agents by proposing \textbf{SlimSearcher}, a training framework driven by a \textbf{Multi-Stage Gating} mechanism. Unlike traditional methods that treat efficiency as a secondary constraint, our approach unifies supervised fine-tuning and reinforcement learning under a single objective. Specifically, it employs a strict correctness gate and uses \textbf{Adaptive Reward Gating} to dynamically optimize computational costs. Extensive experiments on long-horizon benchmarks demonstrate that \textbf{SlimSearcher} reduces average tool-call rounds by 17\% to 58\% without compromising task accuracy. This transforms open-source models into accurate and computationally disciplined agents, offering a scalable approach for developing robust and highly efficient next-generation web agents.

\section*{Limitations}
Despite its superior efficiency and accuracy, \textbf{SlimSearcher} has several limitations that warrant future research. First, our current framework is optimized for text-based reasoning; as web environments become increasingly visual, extending the \textbf{Adaptive Efficiency Anchoring (AEA)} mechanism to handle multimodal redundancy and the high computational cost of processing visual elements remains a critical next step. Second, the effectiveness of our Reinforcement Learning stage is partially tied to the quality of the SFT initialization; in extremely niche domains where the base model fails to discover a single {Minimal Necessary Path}, the AEA mechanism lacks the necessary empirical anchors to bootstrap optimization. Finally, our current implementation employs a uniform weighting scheme for tool calls, which does not account for the varying latency and financial costs associated with different tools in real-world Deep Research deployments. Future work should investigate fine-grained, cost-sensitive tool weighting to better align efficiency rewards with practical operational constraints.

%

\bibliography{main}

\clearpage

\appendix

\section{Implementation Details}
\label{sec:Implementation}

\subsection{Data Preparation}
\label{sec:data_preparation}

To construct a robust and high-signal training environment for \textbf{SlimSearcher}, we curated specialized datasets for both the SFT and RL stages. These high-quality training instances were meticulously curated through iterative rejection sampling and answer consistency verification, drawn from a diverse collection of information-seeking datasets, including Asearcher \cite{gao2025beyond}, TaskCraft \cite{shi2025taskcraft}, WebWalker \cite{wu2025webwalker}, Voyager \cite{he2024webvoyager}, WebShaper \cite{tao2507webshaper}, RedSearcher \cite{chu2026redsearcher}, WebDancer \cite{wu2025webdancer}, and MegaScience \cite{fan2025megascience}. Furthermore, we synthesized a series of additional training instances by building upon the synthetic data generation paradigms introduced by WebSailor \cite{li2025websailor} and WebDancer \cite{li2025websailor}. We present representative cases of our synthesized data in Figure~\ref{fig:synthetic_data_double}. Our data preparation pipeline prioritizes complex, long-horizon web navigation tasks while strictly enforcing efficiency constraints. \textbf{All data will be made publicly available upon acceptance.}

\begin{figure*}[t]
\centering
\framebox{%
\begin{minipage}{0.9\linewidth}
\ttfamily\footnotesize
{\color{blue}Question:}\\
Who propounded the concept of the rule of law in a nation whose diplomatic recognition of a Middle Eastern state---contrary to its closest ally's position---was welcomed by a regional political organization whose legislative body allocates exactly four seats per member, which recently reinstated a country's membership after a decade-long suspension, and whose port serves as a key European stop on a new Arctic shipping route originating from a coastal province which, serving as a junction with both a province ranking second in the nation for its number of sites belonging to a heritage designation that, established in the mid-2010s by an international commission, requires candidate projects to have operated for over a century and whose registry is numerically dominated by entries from a single Asian nation, where the rural tap water penetration rate has exceeded ninety-four percent, and a province where residents have the highest surveyed rate of traditional medicine consumption, even as a separate multi-year analysis shows foreign pharmaceuticals leading sales in its medical institutions, and housing a world-leading battery producer, registers a point-out degree centrality of exactly three in a national study of medical device innovation networks?\\[6pt]
{\color{purple}Answer:}\\
\textbf{Albert Venn Dicey}

\bigskip

{\color{blue}Question:}\\
Between the animation duo---a partnership from a country where, for the first time in over three decades, its two most storied universities failed to place in the top three of a major national ranking while a South China business school was recognized as an academic institution whose diverse partnerships include a third-year student exchange with a university's Weihai campus, a dual-degree program with a Guangdong business school, and a contribution to an international medical consensus statement on aging---and the home nation of a tech giant that established its inaugural AI center in a region whose conflict-related mental health crises are the subject of academic study, while simultaneously becoming a scholarship-funded higher education destination for students from an East Asian nation turning away from Western universities, a country whose national oncological research body is a listed participant in expert reviews by a Swiss-headquartered intergovernmental organization that established a global day for patient safety in the year immediately preceding its declaration of the novel coronavirus as a public health emergency of international concern---whose aesthetic is cited as a key influence on a Czech animator, and the director of a 2015 documentary about autistic youths preparing for a formal dance in an Ohio city, who was the first to win an award?\\[6pt]
{\color{purple}Answer:}\\
\textbf{Brothers Quay}
\end{minipage}
}
\caption{Cases of our synthesized data.}
\label{fig:synthetic_data_double}
\end{figure*}

\subsection{Details of Rejection Sampling}

\label{RS}
To construct a high-quality distillation corpus, we aggregate several public QA datasets alongside partially synthetic data. To ensure the complexity and quality of training signals, we execute each query four times within the \textit{Tongyi-DeepResearch} environment. We specifically retain only samples exhibiting a pass rate within the interval $0 < \mathrm{PR}(q) \le 1$. This filtering criterion targets ``challenging yet solvable'' tasks, effectively excluding trivial queries that do not require deep research and impossible tasks that might introduce noise.
The resulting filtered trajectories serve as the foundation for the \textbf{SFT Stage}. Unlike traditional rejection sampling, which typically selects a successful trajectory at random from the candidate pool, our framework employs \textbf{Pareto-efficient filtration}. Among all correct candidate trajectories $\{\tau_1, \dots, \tau_K\}$ for a given query, we identify and distill the \textit{Minimal Necessary Path} (MNP). We rank these candidates by their joint efficiency score (derived from the weighted tool-call costs and token counts defined in Section~\ref{sec:gating_mechanism}) and select only the top-performing trajectory for fine-tuning.

\subsection{SFT stage}
\label{sec:SFT_stage}

For the SFT stage, we utilized a carefully curated dataset comprising 13,863 training trajectories sourced from Asearcher \cite{gao2025beyond}, TaskCraft \cite{shi2025taskcraft}, WebWalker \cite{wu2025webwalker}, and partially synthetic data filtered via our proposed method (Section~\ref{sec:stage1_sft}). The training is conduct on the ms-swift framework~\cite{zhao2024swiftascalablelightweightinfrastructure}. We implemented a sophisticated two-phase learning rate schedule, beginning with a 5\% warmup period during which the learning rate linearly increases from $1\times10^{-10}$ to $5\times10^{-6}$, followed by a 95\% cosine decay phase that gradually reduces the learning rate from $5\times10^{-6}$ back to $1\times10^{-10}$ to ensure stable convergence.
Training was conducted with a micro-batch size of 1 per GPU and a global batch size of 16, leveraging an advanced \textbf{four-level parallelism strategy}: expert model parallelism (2-way), tensor model parallelism (2-way), pipeline model parallelism (8-way), and context parallelism (2-way). This configuration requires a minimum of 64 GPUs across 8 computational nodes to achieve optimal data parallelism.
This hybrid parallel architecture enables efficient processing of 128K context lengths, with gradient accumulation over 16 steps to reach the target global batch size.

\subsection{RL stage}

To facilitate robust adversarial evolution, we constructed a specialized environment for the RL stage. Our curated training corpus comprises 1,510 high-quality QA pairs sampled from a diverse collection of information-seeking datasets: Voyager \cite{he2024webvoyager}, WebShaper \cite{li2025websailor}, REDSearcher \cite{chu2026redsearcher}, WebDancer \cite{wu2025webdancer}, and MegaScience \cite{fan2025megascience}. The RL in conduct on the RLLM framework~\cite{rllm2025}. We employ the \textbf{GRPO} algorithm~\cite{shao2024deepseekmath} with a training batch size of 64 and a validation batch size of 128. The prompt length is capped at 8,000 tokens, while the maximum response length is set to 120,000 tokens to accommodate long-horizon reasoning trajectories. The actor model is optimized with a learning rate of $1\times10^{-6}$, using a PPO mini-batch size of 32 and dynamic batch sizing with a maximum token length of 16,000 tokens per GPU. 

For rollout, we utilize \textbf{vLLM}~\cite{kwon2023efficient} in asynchronous mode with a tensor model parallel size of 8 and a GPU memory utilization of 0.3. During training, 8 responses are sampled per prompt at temperature 1.0, while validation decoding adopts temperature 0.7, top-$p$ of 0.8, and top-$k$ of 20. Both actor and reference model FSDP configurations enable parameter offloading, and gradient checkpointing is activated to reduce memory overhead. Ulysses sequence parallelism of size 8 is applied to the actor for efficient long-context training.

The entire RL training pipeline is deployed on 64 H800 GPUs across 8 nodes. Each agent is allowed a maximum of 100 environment interaction steps with a trajectory timeout of 7,200 seconds. Training proceeds for up to 10 epochs, with checkpoints saved every 10 steps.

\subsection{Datasets}
\label{sec:dataset}
To comprehensively evaluate the proposed framework, we conduct experiments on four widely adopted web agent benchmarks: XBench-DeepSearch \cite{xbench}, BrowseComp \cite{bc_en}, GAIA \cite{mialon2023gaia}, and HLE \cite{hle}. For the GAIA benchmark, we utilize a subset of 103 text-only queries from the development set. For the HLE benchmark, following the experimental protocol of previous studies \cite{webthinker}, we evaluate the models on a subset of 500 text-only questions.

\subsection{Evaluation Details}

For GAIA and xBench, following the evaluation protocol of \citet{tongyidrs}, we adopt Qwen2.5-72B-Instruct as the judge model. The evaluation prompt remains identical to that used in their work to ensure consistency and comparability. For xBench-DeepSearch, we adopt Gemini-2.5-Flash as the judge model. For BrowseComp, we employ GPT-4o-2024-11-20 as the judge model. For Humanity's Last Exam, we evaluate the 500 text-only questions following~\cite{li2025webthinker}. The evaluation prompt follows the official protocol, with GPT-4o-2024-11-20 serving as the evaluator. The evaluation prompts for all benchmarks are kept consistent with those described in the respective original papers to ensure alignment and reproducibility. The detailed evaluation prompts for each benchmark will be made publicly available on our GitHub repository upon acceptance.

\section{System Prompt}
In this section, we present the complete system prompt used to initialize the \textbf{SlimSearcher} agent. This prompt serves as the foundational instruction set, defining the agent's core identity as a ``Deep Research Assistant.'' It explicitly outlines the available external tools---including web search, page visiting, Python code execution, and file parsing---along with their precise input parameters. Furthermore, it enforces a strict interaction schema, requiring the agent to formulate tool invocations as standardized JSON objects within specific XML tags and to encapsulate its final conclusion within \texttt{<answer>} tags. The exact prompt template is illustrated in Figure~\ref{fig:system_prompt}.

\begin{figure*}[!t]
\begin{tcolorbox}[
  title=System Prompt,
  breakable,
  enhanced,
  fontupper=\small,
  left=2mm, right=2mm, top=1mm, bottom=1mm
]
\setlength{\parskip}{4pt}%
\setlength{\parindent}{0pt}%
You are a deep research assistant. Your core function is to conduct thorough,
multi-source investigations into any topic. You must handle both broad,
open-domain inquiries and queries within specialized academic fields.
For every request, synthesize information from credible, diverse sources
to deliver a comprehensive, accurate, and objective response.
When you have gathered sufficient information and are ready to provide the
definitive response, you must enclose the entire final answer within
\textcolor{black}{\textbf{<answer></answer>}} tags.

\textbf{\# Tools}

You may call one or more functions to assist with the user query.
You are provided with function signatures within
\textcolor{black}{\textbf{<tools></tools>}} XML tags:

\textcolor{black}{\textbf{<tools>}}

\{"type": "function", "function": \{"name":
"\textcolor{red1}{\textbf{search}}",
"description": "Perform Google web searches then returns a string of the top
search results. Accepts multiple queries.",
"parameters": \{"type": "object", "properties": \{"query": \{"type": "array",
"items": \{"type": "string", "description": "The search query."\},
"minItems": 1,
"description": "The list of search queries."\}\},
"required": ["query"]\}\}\}

\{"type": "function", "function": \{"name":
"\textcolor{red1}{\textbf{visit}}",
"description": "Visit webpage(s) and return the summary of the content.",
"parameters": \{"type": "object", "properties": \{%
"url": \{"type": "array", "items": \{"type": "string"\},
"description": "The URL(s) of the webpage(s) to visit."\},
"goal": \{"type": "string",
"description": "The specific information goal for visiting webpage(s)."\}\},
"required": ["url", "goal"]\}\}\}

\{"type": "function", "function": \{"name":
"\textcolor{red1}{\textbf{PythonInterpreter}}",
"description": "Executes Python code in a sandboxed environment.
To use this tool: 1. The \texttt{arguments} JSON object must be empty: \{\}.
2. The Python code must be placed after the JSON block, enclosed within
\textcolor{black}{\textbf{<code></code>}} tags.
Any output MUST be printed using \texttt{print()}.
Example: \textcolor{black}{\textbf{<tool\_call>}}
\{"name": "PythonInterpreter", "arguments": \{\}\}
\textcolor{black}{\textbf{<code>}}
import numpy as np \# Your code here
print(f"The result is: \{np.mean([1,2,3])\}")
\textcolor{black}{\textbf{</code>}}
\textcolor{black}{\textbf{</tool\_call>}}",
"parameters": \{"type": "object", "properties": \{\}, "required": []\}\}\}

\{"type": "function", "function": \{"name":
"\textcolor{red1}{\textbf{\texttt{google\_scholar}}}",
"description": "Leverage Google Scholar to retrieve relevant information
from academic publications. Accepts multiple queries.
This tool will also return results from Google search.",
"parameters": \{"type": "object", "properties": \{"query": \{"type": "array",
"items": \{"type": "string", "description": "The search query."\},
"minItems": 1,
"description": "The list of search queries for Google Scholar."\}\},
"required": ["query"]\}\}\}

\{"type": "function", "function": \{"name":
"\textcolor{red1}{\textbf{parse\_file}}",
"description": "This is a tool that can be used to parse multiple user
uploaded local files such as PDF, DOCX, PPTX, TXT, CSV, XLSX, DOC, ZIP,
MP4, MP3.",
"parameters": \{"type": "object", "properties": \{"files": \{"type": "array",
"items": \{"type": "string"\},
"description": "The file name of the user uploaded local files to be
parsed."\}\}, "required": ["files"]\}\}\}

\textcolor{black}{\textbf{</tools>}}

For each function call, return a json object with function name and arguments
within \textcolor{black}{\textbf{<tool\_call></tool\_call>}} XML tags:

\textcolor{black}{\textbf{<tool\_call>}}

\{"name": <function-name>, "arguments": <args-json-object>\}

\textcolor{black}{\textbf{</tool\_call>}}

Current date:
\end{tcolorbox}
\caption{The complete system prompt used to initialize the SlimSearcher agent.}
\label{fig:system_prompt}
\end{figure*}

\section{Prompt Variations and Analysis of the PromptControl Baseline}
\label{app:prompt_control}

In our main experiments, we evaluated \textbf{PromptControl} as a baseline to determine whether the "efficiency trap" and performative reasoning could be mitigated simply through explicit instructional prompting. While we tested various prompt formulations, empirical results consistently demonstrated that static system instructions are insufficient to override the deeply ingrained verbose behaviors learned during standard SFT.

\subsection{Prompt Variations Tested}
To ensure a rigorous baseline, we experimented with several prompt variations, ranging from soft guidance to strict negative constraints. These directives were appended to the standard system prompt of the agent:

\begin{itemize}
    \item \textbf{Variation 1 (Soft Guidance):} \textit{"Try to be as efficient as possible. Solve the problem using the minimum number of search queries and tool calls."}
    \item \textbf{Variation 2 (Strict Constraint):} \textit{"You are strictly prohibited from making redundant tool calls. Do not use tools if you already know the answer. Keep your reasoning concise."}
    \item \textbf{Variation 3 (Targeted Optimization -- Reported in Main Results):} \textit{"Please note: Minimize the number of tokens you use and the number of tool calls to avoid inefficient searches and visit."}
\end{itemize}

Among these, \textbf{Variation 3} proved to be the most empirically effective in slightly reducing tool usage without immediately collapsing task accuracy. Consequently, this variation was adopted as the ours \textbf{PromptControl} baseline.
Despite employing the optimized prompt, the PromptControl baseline still suffered from severe efficiency degradation on long-horizon tasks (e.g., HLE benchmark). These observations underscore the necessity of the \textbf{SlimSearcher} framework.

\section{Case Study}

\begin{figure*}[!t]
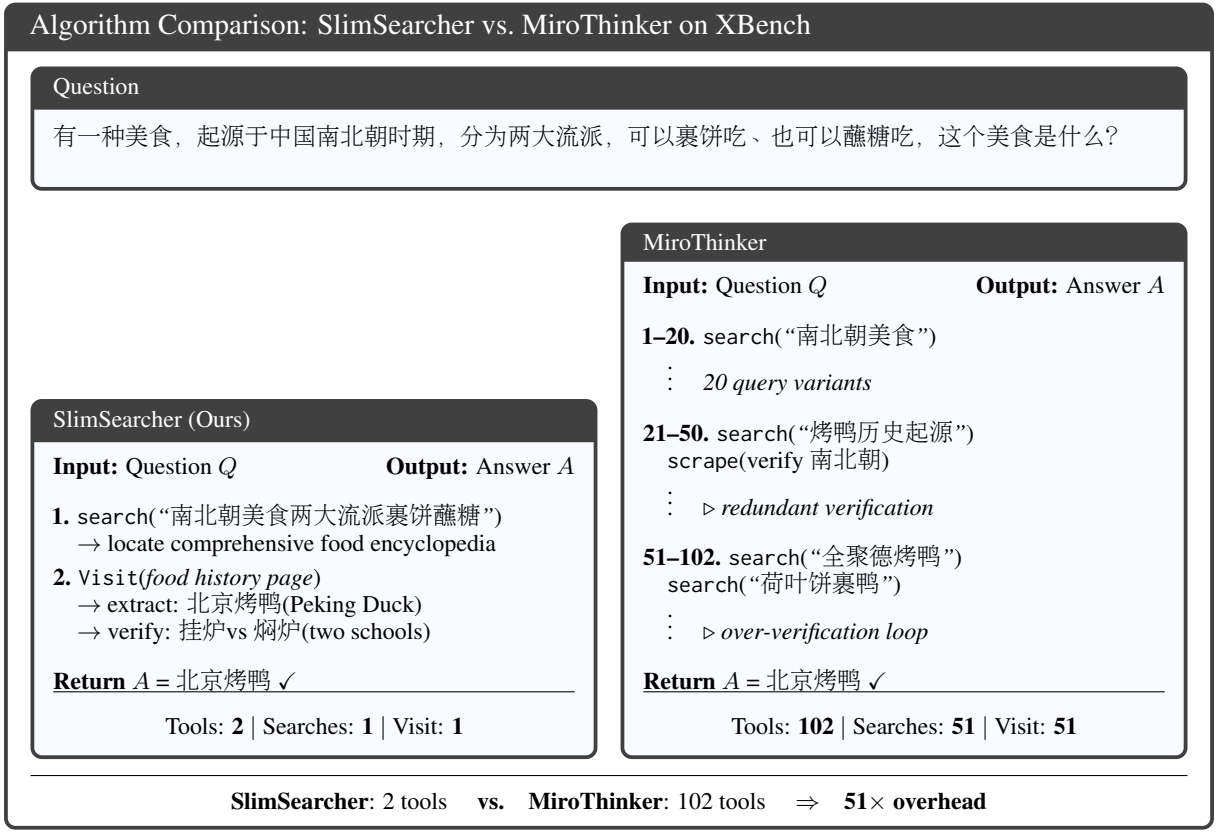

\begin{CJK*}{UTF8}{gbsn} 
\begin{tcolorbox}[
  title=Algorithm Comparison: SlimSearcher vs.\ MiroThinker on XBench ,
  enhanced,
  fontupper=\small,
  left=2mm, right=2mm, top=1mm, bottom=1mm,
  colback=white, 
  colframe=gray!50!black 
]

\begin{tcolorbox}[
  title={\small Question},
  enhanced, fontupper=\small,
  left=1.5mm, right=1.5mm, top=1mm, bottom=1mm,
  colback=lightblue!30 
]
有一种美食，起源于中国南北朝时期，分为两大流派，可以裹饼吃、也可以蘸糖吃，这个美食是什么？\\

\end{tcolorbox}

\medskip
\newpage
\begin{minipage}[t]{0.49\linewidth} 
\begin{tcolorbox}[
  title={\small SlimSearcher (Ours)},
  enhanced, fontupper=\small,
  left=1.5mm, right=1.5mm, top=1mm, bottom=1mm,
  colback=lightblue!30 
]
\setlength{\parskip}{3pt}
\setlength{\parindent}{0pt}

\textbf{Input:} Question $Q$ \hfill \textbf{Output:} Answer $A$

\medskip
\textbf{1.} \texttt{search}(\textit{``南北朝美食 两大流派 裹饼 蘸糖''})\\
\hspace*{1em}$\rightarrow$ locate comprehensive food encyclopedia

\textbf{2.} \texttt{Visit}(\textit{food history page})\\
\hspace*{1em}$\rightarrow$ extract: 北京烤鸭 (Peking Duck)\\
\hspace*{1em}$\rightarrow$ verify: 挂炉 vs 焖炉 (two schools)

\medskip
\textbf{Return} $A$  = {北京烤鸭} \checkmark

\vfill \hrule \medskip 
\centering\footnotesize
Tools: \textbf{2} $\mid$ Searches: \textbf{1} $\mid$ Visit: \textbf{1}
\end{tcolorbox}
\end{minipage}
\hfill
\begin{minipage}[t]{0.49\linewidth}
\begin{tcolorbox}[
  title={\small MiroThinker},
  enhanced, fontupper=\small,
  left=1.5mm, right=1.5mm, top=1mm, bottom=1mm,
  colback=lightblue!30 
]
\setlength{\parskip}{3pt}
\setlength{\parindent}{0pt}

\textbf{Input:} Question $Q$ \hfill \textbf{Output:} Answer $A$

\medskip
\textbf{1--20.} \texttt{search}(\textit{``南北朝美食''})\\
\hspace*{1em}$\vdots$ \quad \textit{20 query variants}

\medskip
\textbf{21--50.} \texttt{search}(\textit{``烤鸭历史起源''})\\
\hspace*{1em}\texttt{scrape}(verify 南北朝)\\
\hspace*{1em}$\vdots$ \quad $\triangleright$ \textit{redundant verification}

\medskip
\textbf{51--102.} \texttt{search}(\textit{``全聚德烤鸭''})\\
\hspace*{1em}\texttt{search}(\textit{``荷叶饼裹鸭''})\\
\hspace*{1em}$\vdots$ \quad $\triangleright$ \textit{over-verification loop}

\medskip
\textbf{Return} $A$ = {北京烤鸭} \checkmark

\vfill \hrule \medskip
\centering\footnotesize
Tools: \textbf{102} $\mid$ Searches: \textbf{51} $\mid$ Visit: \textbf{51}
\end{tcolorbox}
\end{minipage}

\medskip\hrule\medskip
\centering\small
\textbf{SlimSearcher}: 2 tools \quad\textbf{vs.}\quad \textbf{MiroThinker}: 102 tools 
\quad$\Rightarrow$\quad \textbf{51$\times$ overhead}
\end{tcolorbox}
\end{CJK*}
\caption{Comparison of reasoning traces on XBench.}
\label{fig:xbench_task6_comparison}
\end{figure*}

\begin{figure*}[!t]
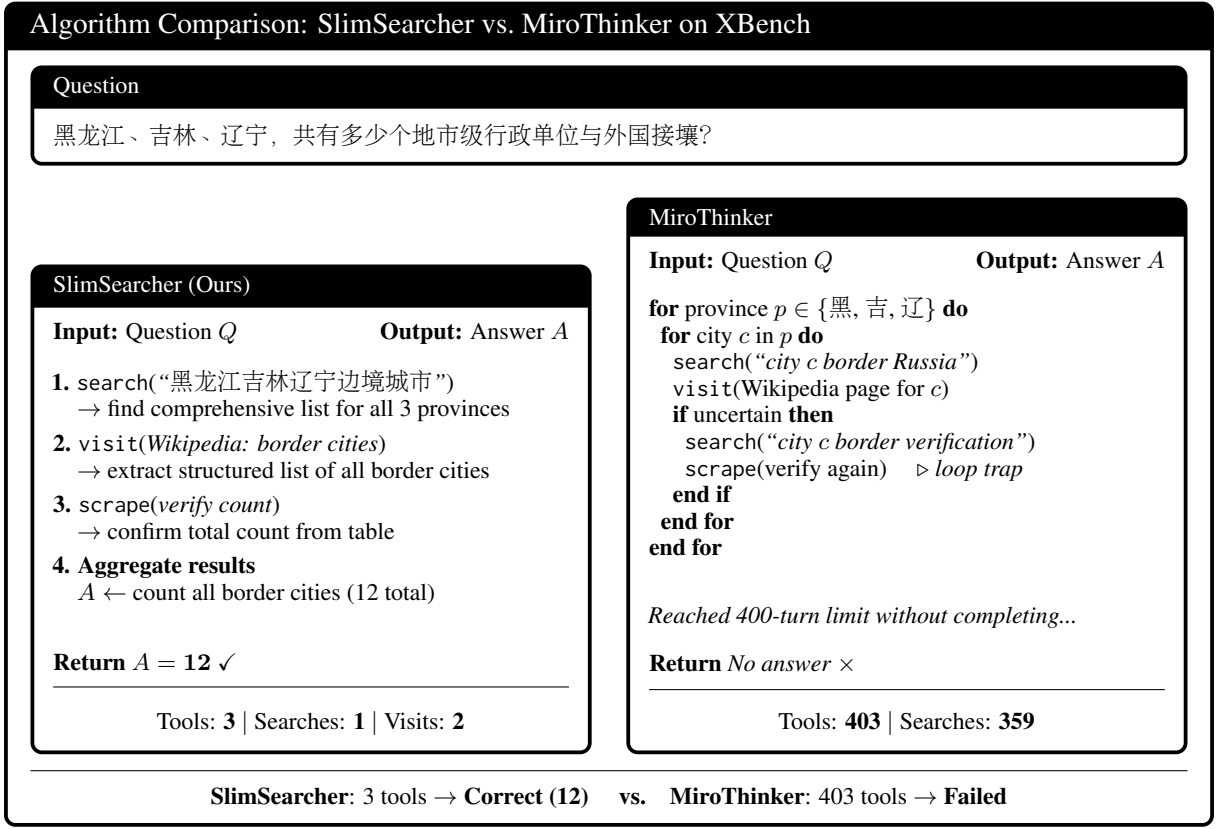

\begin{CJK*}{UTF8}{gbsn}
\begin{tcolorbox}[
  title={Algorithm Comparison: SlimSearcher vs.\ MiroThinker on XBench},
  enhanced, 
  fontupper=\small,
  left=2mm, right=2mm, top=1mm, bottom=1mm,
  colback=white, colframe=black
]

\begin{tcolorbox}[
  title={\small Question},
  enhanced, fontupper=\small,
  left=1.5mm, right=1.5mm, top=1mm, bottom=1mm,
  colback=white, colframe=black
]
黑龙江、吉林、辽宁，共有多少个地市级行政单位与外国接壤？
\end{tcolorbox}

\medskip

\begin{minipage}[t]{0.485\linewidth}
\begin{tcolorbox}[
  title={\small SlimSearcher (Ours)},
  enhanced, fontupper=\small,
  left=1.5mm, right=1.5mm, top=1mm, bottom=1mm,
  colback=white, colframe=black
]
\setlength{\parskip}{3pt}
\setlength{\parindent}{0pt}

\textbf{Input:} Question $Q$ \hfill \textbf{Output:} Answer $A$

\medskip

\textbf{1.} \texttt{search}(\textit{``黑龙江吉林辽宁 边境城市''})\\
\hspace*{1em}$\rightarrow$ find comprehensive list for all 3 provinces

\textbf{2.} \texttt{visit}(\textit{Wikipedia: border cities})\\
\hspace*{1em}$\rightarrow$ extract structured list of all border cities

\textbf{3.} \texttt{scrape}(\textit{verify count})\\
\hspace*{1em}$\rightarrow$ confirm total count from table

\textbf{4.} \textbf{Aggregate results}\\
\hspace*{1em}$A \leftarrow$ count all border cities (12 total)

\vspace{1.5em}
\textbf{Return} $A = \mathbf{12}$ \checkmark

\medskip\hrule\medskip
\centering\footnotesize
Tools: \textbf{3} $\mid$ Searches: \textbf{1} $\mid$ Visits: \textbf{2}
\end{tcolorbox}
\end{minipage}%
\hfill
\begin{minipage}[t]{0.485\linewidth}
\begin{tcolorbox}[
  title={\small MiroThinker},
  enhanced, fontupper=\small,
  left=1.5mm, right=1.5mm, top=1mm, bottom=1mm,
  colback=white, colframe=black
]
\setlength{\parskip}{2pt}
\setlength{\parindent}{0pt}

\textbf{Input:} Question $Q$ \hfill \textbf{Output:} Answer $A$

\medskip

\textbf{for} province $p \in \{$黑, 吉, 辽$\}$ \textbf{do}\\
\hspace*{0.5em}\textbf{for} city $c$ in $p$ \textbf{do}\\
\hspace*{1em}\texttt{search}(\textit{``city $c$ border Russia''})\\
\hspace*{1em}\texttt{visit}(Wikipedia page for $c$)\\
\hspace*{1em}\textbf{if} uncertain \textbf{then}\\
\hspace*{1.5em}\texttt{search}(\textit{``city $c$ border verification''})\\
\hspace*{1.5em}\texttt{scrape}(verify again) \quad $\triangleright$ \textit{loop trap}\\
\hspace*{1em}\textbf{end if}\\
\hspace*{0.5em}\textbf{end for}\\
\textbf{end for}

\vspace{1.5em}
\textit{Reached 400-turn limit without completing...}

\medskip
\textbf{Return} \textit{No answer} \texttimes

\medskip\hrule\medskip
\centering\footnotesize
Tools: \textbf{403} $\mid$ Searches: \textbf{359} 
\end{tcolorbox}
\end{minipage}

\medskip\hrule\medskip
\centering\small
\textbf{SlimSearcher}: 3 tools $\rightarrow$ \textbf{Correct (12)}
\quad\textbf{vs.}\quad
\textbf{MiroThinker}: 403 tools $\rightarrow$ \textbf{Failed}

\end{tcolorbox}
\end{CJK*}
\caption{Comparison of reasoning traces on XBench. SlimSearcher uses a strategic aggregation approach, while MiroThinker falls into an exhaustive verification loop.}
\label{fig:xbench_task11_comparison}
\end{figure*}

To qualitatively evaluate \textbf{SlimSearcher's} capability in circumventing the ``efficiency trap'' inherent in vanilla base models, we conduct a detailed analysis comparing reasoning traces against the baseline agent, MiroThinker. We present three representative examples from GAIA and XBench, visualized in Figures \ref{fig:gaia_detailed_comparison} ,\ref{fig:xbench_task11_comparison} and
\ref{fig:xbench_task6_comparison}.

\vspace{3 cm} 

\begin{figure*}[p]
\begin{tcolorbox}[
  title={SlimSearcher vs.\ MiroThinker},
  enhanced, 
  fontupper=\small,
  left=2mm, right=2mm, top=1mm, bottom=1mm,
  colback=white, colframe=black
]

\begin{tcolorbox}[
  title={\footnotesize Question},
  enhanced, fontupper=\footnotesize,
  left=1.5mm, right=1.5mm, top=0.5mm, bottom=0.5mm,
  colback=white, colframe=black
]
The work referenced in footnote 397 of Federico Lauria's 2014 dissertation
is also the source for the titles of two paintings in the Smithsonian
American Art Museum's collection. What is the absolute difference between
the chapter numbers referenced in the titles of these two paintings?
\end{tcolorbox}

\medskip

\begin{minipage}[t]{0.485\linewidth}
\begin{tcolorbox}[
  title={\footnotesize SlimSearcher (Ours)},
  enhanced, fontupper=\footnotesize,
  left=1.5mm, right=1.5mm, top=0.5mm, bottom=0.5mm,
  colback=white, colframe=black
]
\setlength{\parskip}{1.5pt}\setlength{\parindent}{0pt}

\textbf{Input:} $Q$ \hfill \textbf{Output:} $A$

\smallskip\textbf{Phase 1: Locate Dissertation}\\
\textbf{1.} \texttt{search}(\textit{``Lauria 2014 dissertation''}) $\rightarrow$ thesis PDF\\
\textbf{2.} \texttt{visit}(PDF) $\rightarrow$ download to sandbox

\smallskip\textbf{Phase 2: Extract Footnote}\\
\textbf{3.} \texttt{parse\_pdf}() $\rightarrow$ search ``397''\\
\textbf{4.} \texttt{extract}(fn.\,397) $\rightarrow$ \textit{Hobbes, Leviathan}\\
\textbf{5.} \texttt{verify}() $\rightarrow$ work $=$ \textit{Leviathan} \checkmark

\smallskip\textbf{Phase 3: Find Paintings}\\
\textbf{6.} \texttt{search}(\textit{``SAAM Leviathan paintings''}) $\rightarrow$ collection page\\
\textbf{7.} \texttt{visit}(SAAM) $\rightarrow$ extract list\\
\textbf{8.} \texttt{filter}() $\rightarrow$ $P_1$: Ch.\,15, $P_2$: Ch.\,23

\smallskip\textbf{Phase 4: Calculate}\\
\textbf{9.} $A \leftarrow |15 - 23| = 8$; \texttt{verify}() $\rightarrow$ stop

\vspace{1.5em}
\hrule\smallskip
\centering Tools: \textbf{22} $\mid$ Searches: \textbf{15} \quad
\textbf{Return} $A=\mathbf{8}$ \checkmark
\end{tcolorbox}
\end{minipage}%
\hfill
\begin{minipage}[t]{0.485\linewidth}
\begin{tcolorbox}[
  title={\footnotesize MiroThinker},
  enhanced, fontupper=\footnotesize,
  left=1.5mm, right=1.5mm, top=0.5mm, bottom=0.5mm,
  colback=white, colframe=black
]
\setlength{\parskip}{1.5pt}\setlength{\parindent}{0pt}

\textbf{Input:} $Q$ \hfill \textbf{Output:} $A$

\smallskip\textbf{Loop 1: Search Dissertation (tools 1--45)}\\
\textbf{for} $i=1$ \textbf{to} $8$: \texttt{search}(``Lauria'' variant $i$)\\
\textbf{for} $j=1$ \textbf{to} $12$: \texttt{scrape}(); \textbf{if} uncertain: re-scrape $\triangleright$ \textit{low conf.}\\
\textbf{for} $k=1$ \textbf{to} $10$: \texttt{search}(``footnote 397'' variant $k$)

\smallskip\textbf{Loop 2: Extract Reference (46--95)}\\
\texttt{create\_sandbox}(); \texttt{download\_pdf}()\\
\textbf{for} $m=1$ \textbf{to} $15$: \texttt{parse\_pdf}(method $m$); \texttt{search}(``397'')\\
\textbf{for} $n=1$ \textbf{to} $10$: \texttt{verify}(); \textbf{if} uncertain: re-verify $\triangleright$ \textit{loop trap}

\smallskip\textbf{Loop 3: Find Paintings (96--200)}\\
\textbf{for} $p=1$ \textbf{to} $25$: \texttt{search}(``SAAM Leviathan'' variant $p$)\\
\textbf{for} $q=1$ \textbf{to} $40$: \texttt{scrape}(SAAM); \texttt{extract}(list)\\
\textbf{for} $r=1$ \textbf{to} $20$: \texttt{verify}(titles)

\smallskip\textbf{Loop 4: Final Verification (201--288)}\\
\textbf{for} $s=1$ \textbf{to} $30$: \texttt{search}(verify chapters)\\
\textbf{for} $t=1$ \textbf{to} $30$: \texttt{calculate}($|c_1-c_2|$); \texttt{verify}()\\
\textbf{for} $u=1$ \textbf{to} $28$: \texttt{final\_check}()

\vspace{1.5em}
\hrule\smallskip
\centering Tools: \textbf{288} $\mid$ Searches: \textbf{134} \quad
\textbf{Return} $A=\mathbf{8}$ \checkmark
\end{tcolorbox}
\end{minipage}

\medskip\hrule\medskip
\centering\footnotesize
\textbf{SlimSearcher}: 22 tools, 15 searches \textbf{vs.}
\textbf{MiroThinker}: 288 tools, 134 searches
\end{tcolorbox}
\caption{Reasoning trace comparison on GAIA. SlimSearcher solves in 4 phases with 22 tools; MiroThinker uses $13\times$ more tools across 4 redundant loops yet reaches the same answer.}
\label{fig:gaia_detailed_comparison}
\end{figure*}

\clearpage

\vspace{10 cm} 
\clearpage
    
\end{document}